# STUDY ON PERFORMANCE IMPROVEMENT OF OIL PAINT IMAGE FILTER ALGORITHM USING PARALLEL PATTERN LIBRARY.

Siddhartha Mukherjee[1]

[1]Samsung R&D Institute, India - Bangalore
siddhartha.m@samsung.com / siddhartha2u@gmail.com

*ABSTRACT*

*This paper gives a detailed study on the performance of oil paint image filter algorithm with various parameters applied on an image of RGB model. Oil Paint image processing, being very performance hungry, current research tries to find improvement using parallel pattern library. With increasing kernel-size, the processing time of oil paint image filter algorithm increases exponentially.*

## 1. INTRODUCTION

This document provides an analytical study on the performance of Oil Paint Image Filter Algorithm. There are various popular linear image filters are available. One of them is Oil Paint image effect. This algorithm, being heavy in terms of processing it is investigated in this study. There related studies are detailed in the Section 7.

## 2. BACKGROUND

Modern days, hands are flooded with digital companions, like Digital Camera, Smart Phones and so on. Most of the devices are having built-in camera. People now more keen on using the built-in camera. The computation power of this class of devices is also increasing day by day. The usage of this handheld devices as camera, overshoot the usage of traditional camera in huge number.

The usage of these cameras has become a common fashion of modern life. This has started a new stream of applications. Applications include various categories e.g. image editing, image enhancement, camera extension application and so on. A large group of these applications include applying different kinds of image filters.

Image filters are of different kinds, with respect their nature of processing or mathematical model. Some of the image filters are good in execution-time in comparison with others. The execution time is a very important data, for this category of application development. Oil Paint is one of the very popular linear image filters, which is very heavy in terms of execution.

## 3. INVESTIGATION METHOD

A simple windows application is developed to analyse different types of image filters. The purpose of this windows application is to accept different JPG image files as an input, and apply different kinds of image filters on to it. In the process of applying image filters the application will log the processing time. The overall operation of the application is explained here.

### 3.1. Operational Overview

The application is realised with two primary requirements: input jpg image files and configuration of image filter parameters. To cater requirements, the application is designed with three major components: a user interface, a jpeg image encoder-decoder and image filter algorithm.

The user interface is developed with Microsoft's Win32 programing. The image encoder and decoder component is designed with Windows Imaging Component, provided by Microsoft on windows desktop.

The following flowchart diagram shows the operational overview of the test environment. During this process of testing, the processing time is logged and analysed for the study.

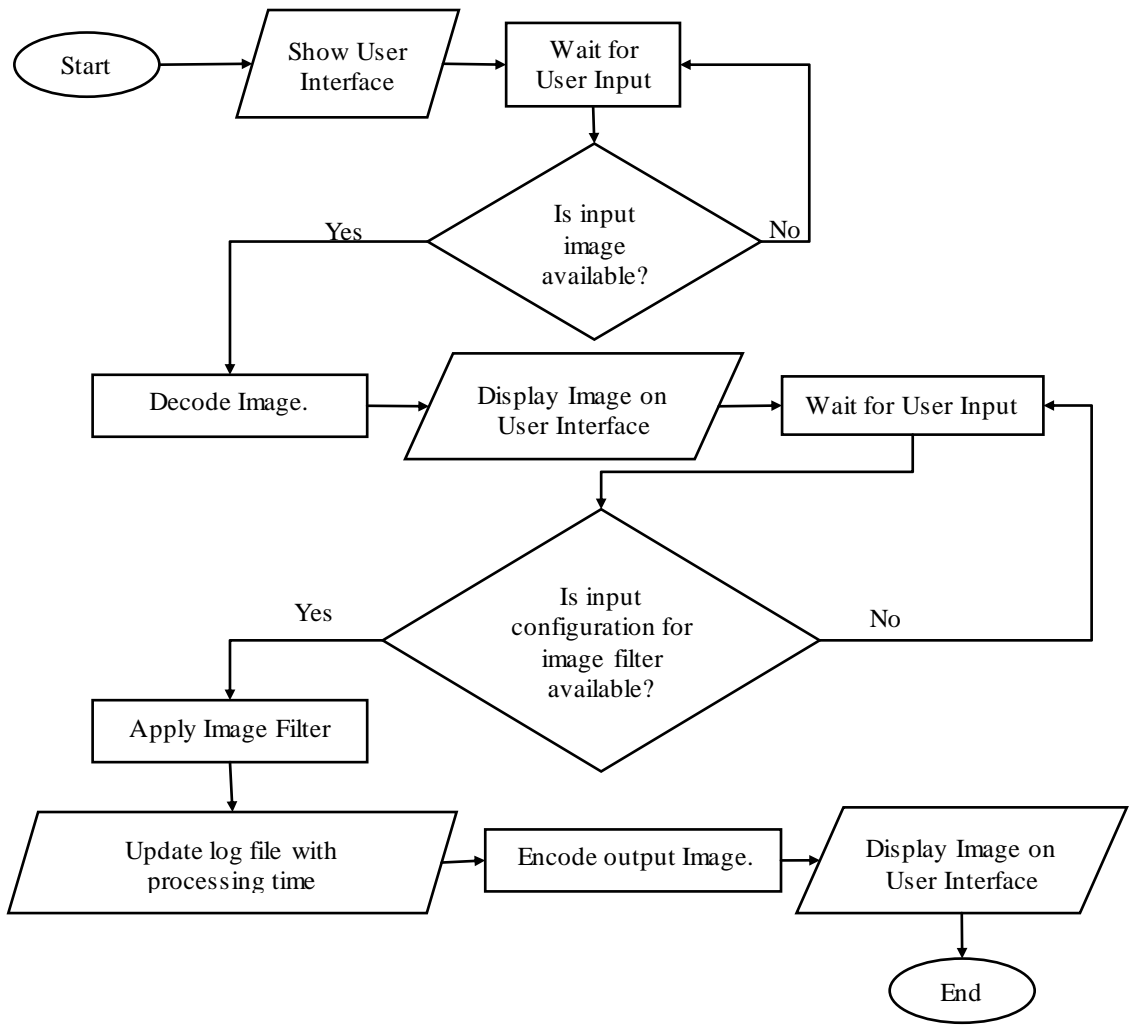

### 3.2. Implementation Overview

Considering the above workflow diagram, main focus of the current study is done with the application of image filter (marked as "Apply Image Filter" operation). Other operations are considered to be well known and do not affect the study. The code snippet below will provide the clear view of implementation. The user interface can be designed in various ways; even this experiment can be performed without a GUI also. That is why the main operational point of interests can be realized with the following way.

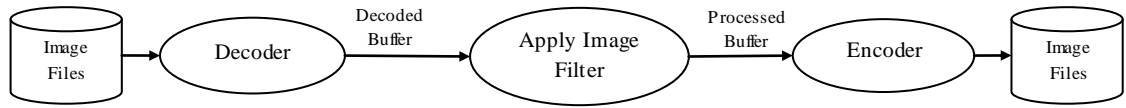

### Decoder

```
/* *******************************************************************************
 * Function Name : Decode
 * Description : The function decodes an image file and gets the decoded buffer.
 *
 * *******************************************************************************/
HRESULT Decode(LPCWSTR imageFilename, PUINT pWidth, PUINT pHeight, PBYTE* ppDecodedBuffer, PUINT pStride,
PUINT pBufferSize, WICPixelFormatGUID*          pWicPixelFormatGUID);
```

One of the possible ways of implementing the decode interface is provided here.

```
HRESULT Decode(LPCWSTR imageFilename, PUINT pWidth, PUINT pHeight, PBYTE* ppDecodedBuffer, PUINT pStride,
PUINT pBufferSize, WICPixelFormatGUID* pWicPixelFormatGUID)
{
        HRESULT                         hr = S_OK;
        UINT                            frameCount = 0;
        IWICImagingFactory              *pFactory = NULL;
        IWICBitmapDecoder               *pBitmapJpgDecoder = NULL;
        IWICBitmapFrameDecode           *pBitmapFrameDecode = NULL;

        do
        {
                /* Create Imaging Factory  */
                BREAK_IF_FAILED(CoCreateInstance( CLSID_WICImagingFactory, NULL, CLSCTX_INPROC_SERVER,
                                IID_IWICImagingFactory, (LPVOID*)&pFactory))

                /*  Create Imaging Decoder for JPG File */
                BREAK_IF_FAILED(pFactory->CreateDecoderFromFilename(imageFilename, NULL, GENERIC_READ,
                                            WICDecodeMetadataCacheOnDemand, &pBitmapJpgDecoder))

                /* Get decoded frame & its related information from Imaging Decoder for JPG File */
                BREAK_IF_FAILED(pBitmapJpgDecoder->GetFrameCount(&frameCount))

                BREAK_IF_FAILED(pBitmapJpgDecoder->GetFrame(0, &pBitmapFrameDecode))

                /* Get Width and Height of the Frame */
                BREAK_IF_FAILED(pBitmapFrameDecode->GetSize(pWidth, pHeight))

                /* Get Pixel format and accordingly allocate memory for decoded frame */
                BREAK_IF_FAILED(pBitmapFrameDecode->GetPixelFormat(pWicPixelFormatGUID))

                *ppDecodedBuffer = allocateBuffer(pWicPixelFormatGUID, *pWidth, *pHeight,
                                            pBufferSize, pStride))
                if(*ppDecodedBuffer == NULL) break;

                /* Get decoded frame  */
                BREAK_IF_FAILED(pBitmapFrameDecode->CopyPixels(NULL, *pStride,
                                            *pBufferSize, *ppDecodedBuffer))

        }while(false);

        if(NULL != pBitmapFrameDecode)      pBitmapFrameDecode->Release();
        if(NULL != pBitmapJpgDecoder)       pBitmapJpgDecoder->Release();
        if(NULL != pFactory)                pFactory->Release();

        return hr;
}
```

**Encoder**

The interface for encoding is exposed as shown here.

```
/* *****************************************************************************
 * Function Name : Encode
 *
 * Description : The function encodes an deocoded buffer into an image file.
 *
 * *****************************************************************************/
HRESULT Encode(LPCWSTR outFilename, UINT imageWidth, UINT imageHeight, PBYTE pDecodedBuffer, UINT cbStride,
UINT cbBbufferSize, WICPixelFormatGUID*      pWicPixelFormatGUID);
```

One of the possible ways of implementing the encode interface is provided here.

```
HRESULT Encode(LPCWSTR outFilename, UINT imageWidth, UINT imageHeight, PBYTE pDecodedBuffer, UINT cbStride, UINT
               cbBbufferSize, WICPixelFormatGUID*   pWicPixelFormatGUID)
{
        HRESULT                         hr = S_OK;
        UINT                            frameCount = 0;
        IWICImagingFactory              *pFactory = NULL;
        IWICBitmapEncoder               *pBitmapJpgEncoder = NULL;
        IWICBitmapFrameEncode           *pBitmapFrameEncode = NULL;
        IWICStream                      *pJpgFileStream = NULL;

        do
        {
                /* Create Imaging Factory  */
                BREAK_IF_FAILED(CoCreateInstance(CLSID_WICImagingFactory, NULL, CLSCTX_INPROC_SERVER,
                                                 IID_IWICImagingFactory, (LPVOID*)&pFactory))

                /* Create & Initialize Stream for an output JPG file */
                BREAK_IF_FAILED(pFactory->CreateStream(&pJpgFileStream))

                BREAK_IF_FAILED(pJpgFileStream->InitializeFromFilename(outFilename, GENERIC_WRITE))

                /* Create & Initialize Imaging Encoder  */
                BREAK_IF_FAILED(pFactory->CreateEncoder(GUID_ContainerFormatJpeg,
                                                 &GUID_VendorMicrosoft,
                                                         &pBitmapJpgEncoder))

                /* Initialize a JPG Encoder */
                BREAK_IF_FAILED(pBitmapJpgEncoder->Initialize(pJpgFileStream, WICBitmapEncoderNoCache))

                /* Create & initialize a JPG Encoded frame */
                BREAK_IF_FAILED(pBitmapJpgEncoder->CreateNewFrame(&pBitmapFrameEncode, NULL))
                BREAK_IF_FAILED(pBitmapFrameEncode->Initialize(NULL))

                /* Update the pixel information */
                BREAK_IF_FAILED(pBitmapFrameEncode->SetPixelFormat(pWicPixelFormatGUID))
                BREAK_IF_FAILED(pBitmapFrameEncode->SetSize(imageWidth, imageHeight))
                BREAK_IF_FAILED(pBitmapFrameEncode->WritePixels(imageHeight, cbStride,
                                                         cbBbufferSize, pDecodedBuffer))

                BREAK_IF_FAILED(pBitmapFrameEncode->Commit())
                BREAK_IF_FAILED(pBitmapJpgEncoder->Commit())

        }while(false);

        if(NULL != pJpgFileStream)          pJpgFileStream->Release();
        if(NULL != pBitmapFrameEncode)      pBitmapFrameEncode->Release();
        if(NULL != pBitmapJpgEncoder)       pBitmapJpgEncoder->Release();
        if(NULL != pFactory)                pFactory->Release();

        return hr;
}
```

```
/* *****************************************************************************
 * Utility Macros
 * *****************************************************************************/
#define BREAK_IF_FAILED(X)   hr = X; \
                      if(FAILED(hr)) { break; } \
```

**Application of Image Filter**

The image processing algorithm is the subject of study in current experiment. Details of the algorithms are explained later sections. Following code snippet will explain how the performances for any simple filter (e.g. oil paint) captured for study. Similar approach is followed all image filters.

```c
HRESULT ApplyOilPaintOnFile (LPCWSTR inImageFile, LPCWSTR outImageFile)
{
        HRESULT             hr = S_OK;
        PBYTE               pDecodedBuffer = NULL;
        PBYTE               pOutputBuffer = NULL;
        UINT                decodedBufferLen = 0;
        UINT                inImageWidth = 0;
        UINT                inImageHeight = 0;
        UINT                cbStride = 0;
        WICPixelFormatGUID  wicPixelFormatGUID;
        DWORD               dTimeStart = 0;
        DWORD               dTimeDecode = 0;
        DWORD               dTimeProcess = 0;
        DWORD               dTimeEncode = 0;
        char                sMessage[256] = {0};

        do
        {
                /* --------- Decode. --------- */
                dTimeStart = GetTickCount();

                BREAK_IF_FAILED(Decode( inImageFile, &inImageWidth, &inImageHeight, &pDecodedBuffer,
                                    &cbStride, &decodedBufferLen, &wicPixelFormatGUID))

                dTimeDecode = GetTickCount() - dTimeStart;

                /* Allocate Memory for output. */
                pOutputBuffer = (PBYTE)calloc(sizeof(BYTE), decodedBufferLen);
                if(NULL == pOutputBuffer)
                        break;

                /* ------------ Process Image Filter ------------ */
                dTimeStart = GetTickCount();

                BREAK_IF_FAILED(ApplyOilPaintOnBuffer(pDecodedBuffer,
                                                inImageWidth, inImageHeight, pOutputBuffer))

                dTimeProcess = GetTickCount() - dTimeStart;

                /* --------- Encode ---------  */
                dTimeStart = GetTickCount();

                BREAK_IF_FAILED(Encode( outImageFile, inImageWidth, inImageHeight, pOutputBuffer,
                                    cbStride, decodedBufferLen, &wicPixelFormatGUID))

                dTimeEncode = GetTickCount() - dTimeStart;

                sprintf(sMessage,
                "Grey Scale : Width=%d, Height=%d, Time(Decode)=%lu Time(Process)=%lu Time(Encode)=%lu\r\n",
                 inImageWidth, inImageHeight, dTimeDecode, dTimeProcess, dTimeEncode);

                Log(sMessage);

        }while(false);

        if(NULL != pDecodedBuffer)  free(pDecodedBuffer);
        if(NULL != pOutputBuffer)   free(pOutputBuffer);

        return hr;
}
```

For measuring the time taken for processing, well known standard windows API GetTickCount is used. GetTickCount retrieves the number of milliseconds that have elapsed since the system was started.

## 4. OIL PAINT IMAGE FILTER IN RGB COLOUR MODEL

During this study, the input images are considered to be in RGB model. In this model, an image consists of two dimensional arrays of pixels. Each pixel of a 2D array contains data of red, green and blue colour channel respectively.

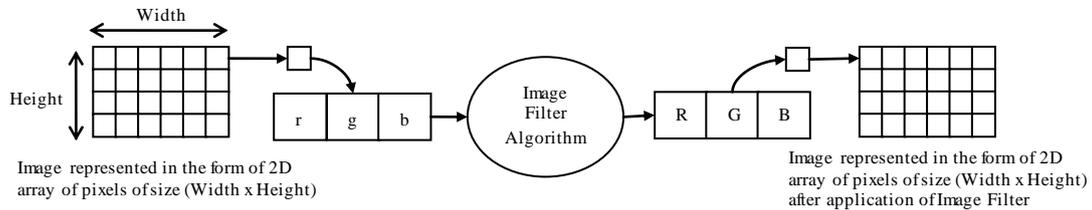

The Image Filters are basically algorithm for changing the values of Red, Green and Blue component of a pixel to a certain value.

There are various kinds of Image Filters, available. One of the categories of image filter is linear image filters. For processing one pixel its neighbouring pixels is accessed, in linear image filter. Depending upon the amount of access to neighbouring pixels, the performance of linear filters is affected.

As a part of our analysis we have considered Oil Paint image filter, which is popular but process hungry.

**Histogram based algorithm for Oil Paint**

For each pixel, it is done in this way: for pixel at position (x, y), find the most frequently occurring intensity value in its neighbourhood. And set it as the new colour value at position (x, y).

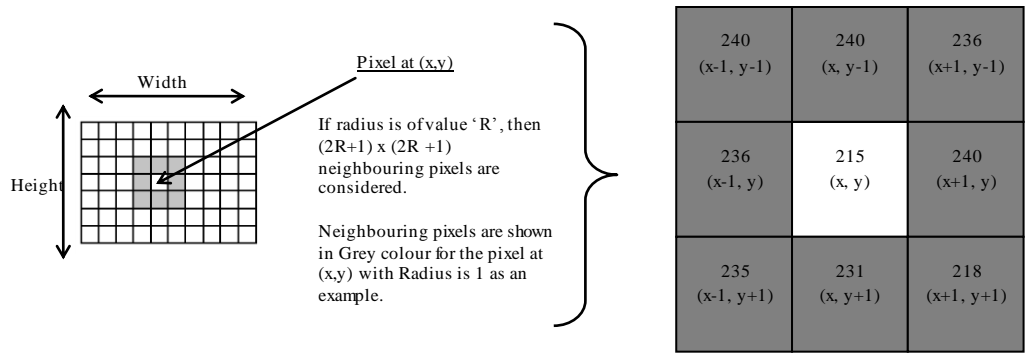

1) The right side provides the larger and clear picture of the neighbouring pixels or Radius 1, with respect to pixel at (x, y). The intensities of the respective pixels are also provided (as an example).
2) The pixels at (x-1, y-1), (x, y-1), (x+1, y) have the maximum occurring intensity i.e. 240.
3) The each colour channel of the pixel at (x, y) is set with an average of each colour channel of 3 pixels [(x-1, y-1), (x, y-1), (x+1, y)].

The interface for the oil paint algorithm is exposed as follows.

```
/* ****************************************************************************
 * Function Name : ApplyOilPaintOnBuffer
 * Description : Apply oil paint effect on decoded buffer.
 *
 * ****************************************************************************/
HRESULT ApplyOilPaintOnBuffer(PBYTE pInBuffer, UINT width, UINT height, const UINT intensity_level, const int radius, PBYTE pOutBuffer);
```

Generally bigger images will be captured with higher resolution cameras. Here the radius also needs to be of higher value to create better oil paint image effect. And this creates more performance bottleneck.

One of the possible ways of implementing the interface is as follows.

```c
HRESULT ApplyOilPaintOnBuffer(PBYTE pInBuffer, UINT width, UINT height, const UINT intensity_level,
const int radius, PBYTE pOutBuffer)
{
        int                             index = 0;
        int                             intensity_count[255] = {0};
        int                             sumR[255] = {0};
        int                             sumG[255] = {0};
        int                             sumB[255] = {0};
        int                             current_intensity = 0;
        int                             row,col, x,y;
        BYTE                            r,g,b;
        int                             curMax = 0;
        int                             maxIndex = 0;

        if(NULL == pInBuffer || NULL == pOutBuffer)
                return E_FAIL;

        for(col = radius; col < (height - radius); col++) {
                for(row = radius; row < (width - radius); row++) {
                        memset(&intensity_count[0], 0, ARRAYSIZE(intensity_count));
                        memset(&sumR[0], 0, ARRAYSIZE(sumR));
                        memset(&sumG[0], 0, ARRAYSIZE(sumG));
                        memset(&sumB[0], 0, ARRAYSIZE(sumB));

                        /* Calculate the highest intensity Neighbouring Pixels. */
                        for(y = -radius; y <= radius; y++) {
                                for(x = -radius; x <= radius; x++) {
                                    index = ((col + y) * width * 3) + ((row + x) * 3);

                                    r = pInBuffer[index + 0];
                                    g = pInBuffer[index + 1];
                                    b = pInBuffer[index + 2];

                                    current_intensity = ((r + g + b) * intensity_level/3.0)/255;
                                    intensity_count[current_intensity]++;

                                    sumR[current_intensity] += r;
                                    sumG[current_intensity] += g;
                                    sumB[current_intensity] += b;
                                }
                        }

                        index = (col * width * 3) + (row * 3);

                    /* The highest intensity neighbouring pixels are averaged out to get the exact color. */
                    maxIndex = 0;
                    curMax = intensity_count[maxIndex];

                    for( int i = 0; i < intensity_level; i++ )  {
                       if( intensity_count[i] > curMax )   {
                          curMax = intensity_count[i];
                          maxIndex = i;
                       }
                    }

                if(curMax > 0) {
                    pOutBuffer[index + 0] = sumR[maxIndex]/curMax;
                    pOutBuffer[index + 1] = sumG[maxIndex]/curMax;
                    pOutBuffer[index + 2] = sumB[maxIndex]/curMax;
                }
            }
        }

        return S_OK;
}
```

## Experimental Results

The experimental is conducted with images of different size and application of oil paint with different radius. The following data shows the time of execution with different parameters.

| Size | Radius | Time |
|---|---|---|
| VGA(640x480) | 2 | 218 |
| VGA(640x480) | 4 | 531 |
| VGA(640x480) | 6 | 1046 |
| VGA(640x480) | 8 | 1685 |
| SVGA(800x600) | 2 | 297 |
| SVGA(800x600) | 4 | 826 |
| SVGA(800x600) | 6 | 1606 |
| SVGA(800x600) | 8 | 2652 |
| XGA(1024x768) | 2 | 499 |
| XGA(1024x768) | 4 | 1326 |
| XGA(1024x768) | 6 | 2621 |
| XGA(1024x768) | 8 | 4383 |
| FHD(1920x1080) | 2 | 1466 |
| FHD(1920x1080) | 4 | 3526 |
| FHD(1920x1080) | 6 | 7020 |
| FHD(1920x1080) | 8 | 11716 |
| WQXGA(2560x1600) | 2 | 2559 |
| WQXGA(2560x1600) | 4 | 6973 |
| WQXGA(2560x1600) | 6 | 14008 |
| WQXGA(2560x1600) | 8 | 23229 |

| System | Details |
|---|---|
| Processor | Intel® Core™ i7-3630QM CPU @ 2.40 GHz, 2.40 GHz |
| Operating System | Windows 7 Enterprise, 64 bit Operating System. |
| RAM | 8.00GB |

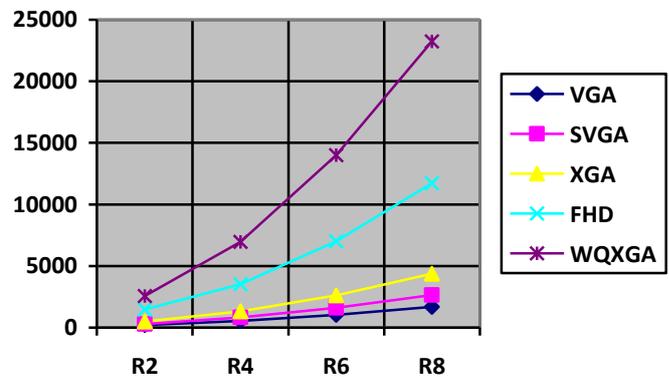

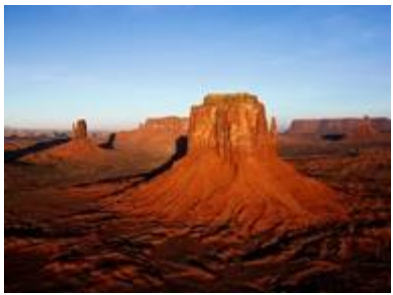
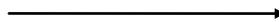
After application of Oil Paint Image Filter
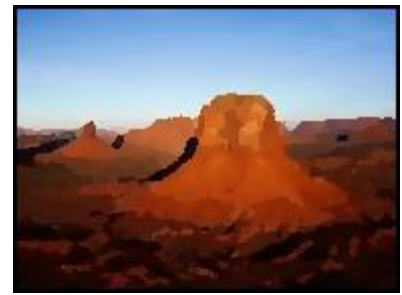

In due course of our investigation, we have observed that the performance of oil paint image filter increases in greater degree with increasing width, height and radius (i.e. usage of neighbouring pixel).

More importantly, we have observed most of the high resolution images are captured by more and more and power camera (i.e. either in high end digital camera or high end handheld devices). For these kinds of higher resolution photos, as the resolution of the image increases, the radius parameter needs to be increased to generate Oil Paint effect of an acceptable quality.

# 5. OIL PAINT IMAGE FILTER BY MICROSOFT PARALLEL PATTERNS LIBRARY

We have observed, in our previous section of investigation, which time increases with higher degree with increasing width, height and radius. So we tried to improve the oil paint algorithm by using Microsoft Parallel Patterns Library. We have kept the same interface for Oil Paint algorithm; only we differentiated in the implementation. Following code snippet will provide clear picture of the implementation using Microsoft PPL.

```cpp
HRESULT ApplyOilPaintOnBuffer(PBYTE pInBuffer, UINT width, UINT height, const UINT intensity_level, const int radius, PBYTE pOutBuffer)
{
        int tStart = radius;
        int tEnd =(height - radius);

        if(NULL == pInBuffer || NULL == pOutBuffer)
                    return E_FAIL;

        parallel_for(tStart, tEnd, [&pInBuffer, &width, &height, &intensity_level, &radius, &pOutBuffer]
        (int col){
                int                     index = 0;
                int                     intensity_count[255] = {0};
                int                     sumR[255] = {0};
                int                     sumG[255] = {0};
                int                     sumB[255] = {0};
                int                     current_intensity = 0;
                int                     row,x,y;
                BYTE                    r,g,b;
                int                     curMax = 0;
                int                     maxIndex = 0;

                for(row = radius; row < (width - radius); row++)
                {
                    /* This portion of the code remains same, as mentioned above */
                }

        });

        return S_OK;
}
```

## Experimental Results

The experiment is conducted with same set of images, used for the experiment, mentioned in the section above. We have also obtained same quality of output with and better performance.

| Size | Radius | Time |
|---|---|---|
| VGA(640x480) | 2 | 94 |
| VGA(640x480) | 4 | 156 |
| VGA(640x480) | 6 | 281 |
| VGA(640x480) | 8 | 483 |
| SVGA(800x600) | 2 | 78 |
| SVGA(800x600) | 4 | 234 |
| SVGA(800x600) | 6 | 452 |
| SVGA(800x600) | 8 | 734 |
| XGA(1024x768) | 2 | 140 |
| XGA(1024x768) | 4 | 375 |
| XGA(1024x768) | 6 | 733 |
| XGA(1024x768) | 8 | 1248 |
| FHD(1920x1080) | 2 | 343 |
| FHD(1920x1080) | 4 | 967 |
| FHD(1920x1080) | 6 | 1935 |
| FHD(1920x1080) | 8 | 3261 |
| WQXGA(2560x1600) | 2 | 686 |
| WQXGA(2560x1600) | 4 | 1872 |
| WQXGA(2560x1600) | 6 | 3915 |
| WQXGA(2560x1600) | 8 | 6490 |

| System | Details |
|---|---|
| Processor | Intel® Core™ i7-3630QM CPU @ 2.40 GHz, 2.40 GHz |
| Operating System | Windows 7 Enterprise, 64 bit Operating System. |
| RAM | 8.00GB |

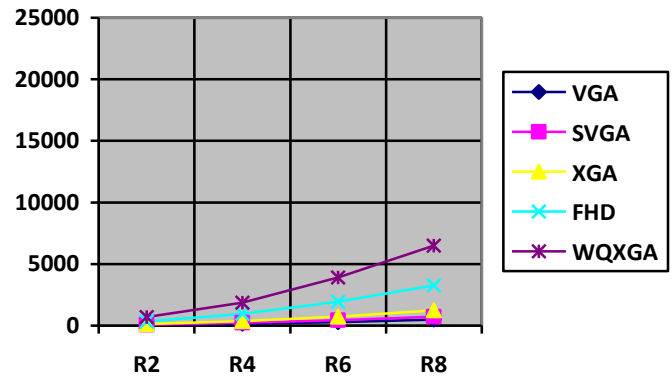

## 6. COMPARATIVE ANALYSIS OF BOTH APPROACHES

The improvement of the performance in terms of percentage is deduced as $[100 * (T1 - T2)/ t1]$, where T1 is time required for processing by $1^{st}$ approach and T2 is the time required for processing time by latest approach.

| Size | Radius | T1 | T2 | Improvement (%) |
|---|---|---|---|---|
| VGA(640x480) | 2 | 218 | 94 | 56.88073394 |
| VGA(640x480) | 4 | 531 | 156 | 70.62146893 |
| VGA(640x480) | 6 | 1046 | 281 | 73.13575526 |
| VGA(640x480) | 8 | 1685 | 483 | 71.33531157 |
| SVGA(800x600) | 2 | 297 | 78 | 73.73737374 |
| SVGA(800x600) | 4 | 826 | 234 | 71.67070218 |
| SVGA(800x600) | 6 | 1606 | 452 | 71.85554172 |
| SVGA(800x600) | 8 | 2652 | 734 | 72.32277526 |
| XGA(1024x768) | 2 | 499 | 140 | 71.94388778 |
| XGA(1024x768) | 4 | 1326 | 375 | 71.71945701 |
| XGA(1024x768) | 6 | 2621 | 733 | 72.03357497 |
| XGA(1024x768) | 8 | 4383 | 1248 | 71.52635181 |
| FHD(1920x1080) | 2 | 1466 | 343 | 76.60300136 |
| FHD(1920x1080) | 4 | 3526 | 967 | 72.57515598 |
| FHD(1920x1080) | 6 | 7020 | 1935 | 72.43589744 |
| FHD(1920x1080) | 8 | 11716 | 3261 | 72.16626835 |
| WQXGA(2560x1600) | 2 | 2559 | 686 | 73.19265338 |
| WQXGA(2560x1600) | 4 | 6973 | 1872 | 73.15359243 |
| WQXGA(2560x1600) | 6 | 14008 | 3915 | 72.05168475 |
| WQXGA(2560x1600) | 8 | 23229 | 6490 | 72.06078609 |

## 7. REFERENCES

From reference [3] the histogram based analysis has been studied. The reference [3] provides the algorithm for the implementation of oil pain image filter algorithm. The algorithm (mentioned in reference [3], section 'Oil-paint Effect') is implemented, as explained in the section 4 of this paper. The achieved performance of the algorithm is examined and captured in the section 4 (sub-section: Experimental Result) here. The result shows high growth of the processing time with respect to kernel-size. Reference [4] is another reference, where algorithm similar reference [3] is proposed for implementation. The reference [1] and [2] are used for way of analysis and follow the broadened scope in this arena of image processing. Reference [5] also proposes algorithm which are similar in nature with reference [3]. So we can clearly depict algorithms similar to reference [3] and [5], will face similar performance problem.

## 8. CONCLUSIONS

As mentioned in section 4 & 7, I have obtained result, which depicts huge growth in processing time with respect to the increase in kernel size of oil paint image filter. There are various approaches have been proposed for the betterment of processing performance of the image filter algorithms. The parallel pattern library is a Microsoft library designed for the use by native C++ developers, which provides features of multicore programming. The current paper conducts study on improving oil paint image filter algorithm using the Microsoft technology.

By comparing results, as shown in section 6 I conclude that by using Microsoft Parallel Pattern library 71.6% (average) performance improvement can be achieved for Oil Paint Algorithm. This study is applicable for similar class of image filter algorithms as well.

There are various similar image filter algorithm, where processing of a single pixel depends on the values of its neighbouring pixels. In this respect, if the larger neighbouring pixels are accessed, there are performance issues. The approach mentioned in this paper can be referred for similar issues.

In future, more well-known or new techniques in conjunction with the current idea can be used for betterment. Not only in image processing in other dimensions of signal processing as well similar approach can be tried.

## ACKNOWLEDGEMENTS

I would like to thank my organization to provide me the opportunity for conducting this research!

**Authors**

**Siddhartha Mukherjee** is a B.Tech (Computer Science and Engineering) from *RCC Institute of Information Technology, Kolkata*. Siddhartha is currently working as a Technical Manager in Samsung R&D Institute, India- Bangalore. Siddhartha has almost 10 years of working experience in software development. He has previously worked with Wipro Technologies. He has been contributing for various technical papers & innovations at different forums in Wipro and Samsung. His main area of work is mobile application developments.

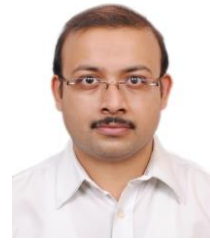